\documentclass[conference]{IEEEtran}
\IEEEoverridecommandlockouts
\usepackage{cite}
\usepackage{amsmath,amssymb,amsfonts}
\usepackage{algorithmic}
\usepackage{graphicx}
\usepackage{booktabs}
\usepackage{threeparttable}
\usepackage{textcomp}
\usepackage{xcolor}
\usepackage{verbatim}
\def\BibTeX{{\rm B\kern-.05em{\sc i\kern-.025em b}\kern-.08em
    T\kern-.1667em\lower.7ex\hbox{E}\kern-.125emX}}
\begin{document}

\title{Automatic Micro-sleep Detection under Car-driving Simulation Environment using Night-sleep EEG \\

\author{\IEEEauthorblockN{Young-Seok Kweon}
\IEEEauthorblockA{\textit{Dept. Brain and Cognitive Engineering} \\
\textit{Korea University}\\
Seoul, Republic of Korea \\
youngseokkweon@korea.ac.kr} \\

\IEEEauthorblockN{Gi-Hwan Shin}
\IEEEauthorblockA{\textit{Dept. Brain and Cognitive Engineering} \\
\textit{Korea University}\\
Seoul, Republic of Korea \\
gh\_shin@korea.ac.kr} 

\and

\IEEEauthorblockN{Heon-Gyu Kwak}
\IEEEauthorblockA{\textit{Dept. Artificial Intelligence} \\
\textit{Korea University}\\
Seoul, Republic of Korea \\
hg\_kwak@korea.ac.kr} \\

\IEEEauthorblockN{Minji Lee}
\IEEEauthorblockA{\textit{Dept. Brain and Cognitive Engineering} \\
\textit{Korea University}\\
Seoul, Republic of Korea \\
minjilee@korea.ac.kr}
}

\thanks{20xx IEEE. Personal use of this material is permitted. Permission
from IEEE must be obtained for all other uses, in any current or future media, including reprinting/republishing this material for advertising or promotional purposes, creating new collective works, for resale or redistribution to servers or lists, or reuse of any copyrighted component of this work in other works.

This work was partly supported by Institute for Information $\&$ communications Technology Planning $\&$ Evaluation (IITP) grant funded by the Korea government (MSIT) (No. 2017-0-00451, Development of BCI based Brain and Cognitive Computing Technology for Recognizing User’s Intentions using Deep Learning) and (No. 2019-0-00079, Artificial Intelligence Graduate School Program (Korea University)).}}
\maketitle

\begin{abstract}
A micro-sleep is a short sleep that lasts from 1 to 30 secs. Its detection during driving is crucial to prevent accidents that could claim a lot of people's lives. Electroencephalogram (EEG) is suitable to detect micro-sleep because EEG was associated with consciousness and sleep. Deep learning showed great performance in recognizing brain states, but sufficient data should be needed. However, collecting micro-sleep data during driving is inefficient and has a high risk of obtaining poor data quality due to noisy driving situations. Night-sleep data at home is easier to collect than micro-sleep data during driving. Therefore, we proposed a deep learning approach using night-sleep EEG to improve the performance of micro-sleep detection. We pre-trained the U-Net to classify the 5-class sleep stages using night-sleep EEG and used the sleep stages estimated by the U-Net to detect micro-sleep during driving. This improved micro-sleep detection performance by about 30\% compared to the traditional approach. Our approach was based on the hypothesis that micro-sleep corresponds to the early stage of non-rapid eye movement (NREM) sleep. We analyzed EEG distribution during night-sleep and micro-sleep and found that micro-sleep has a similar distribution to NREM sleep. Our results provide the possibility of similarity between micro-sleep and the early stage of NREM sleep and help prevent micro-sleep during driving.
\end{abstract}


\section{Introduction}
A micro-sleep is a temporary sleep state which holds for a short period from 1 sec to 30 secs \cite{Buriro2018}. It occurs when there are sudden shifts between states of wakefulness and sleep. Monotonous tasks, including driving a car and watching an empty computer screen, induce a micro-sleep \cite{Poudel2009}. In a driving situation, reaction time and task-performance can be negatively affected by a micro-sleep similar or even more severe than driving with an illegal blood alcohol concentration \cite{Williamson2000}. Therefore, micro-sleep during driving can cause serious accidents, which may claim many people's lives.

To prevent the micro-sleep, it is essential to detect and wake up in this state. Monitoring brain signals is reasonable to detect a micro-sleep because consciousness and sleep are considerably associated with a brain \cite{Suk2011, Simon1956, Lee2017, Yeom2017, Lee2019}. Electroencephalogram (EEG) is a brain monitoring method using electrical potential differences in scalp \cite{Baillet2001, kam2013, kim2014, Lee2015, Chen2016}. Since portability and cost-efficiency of EEG are better than other brain monitoring methods, including functional magnetic resonance imaging (fMRI) \cite{Alonso2012, Kwak2017, Kim2018, Kwak2020}, recent studies have been used EEG to detect micro-sleep \cite{Buriro2018, Liang2019, Skorucak2020}. They showed high performance of micro-sleep detection using machine learning, including linear discriminant analysis \cite{Buriro2018}, logistic regression \cite{Liang2019}, and support vector machine \cite{Skorucak2020}. However, their methods used multi-channel EEG and eye movement information, which are limited to obtain data in real driving situations.

The deep learning algorithms have shown the great performance of various tasks \cite{Park2016}, such as sleep staging classification using single-channel EEG, including DeepSleepNet (DSN) \cite{Supratak2017} and U-time \cite{Perslev2019}. DSN was composed of a two-stream convolutional neural network (CNN) and long short-term memory (LSTM) to divide sleep EEG into frequency, temporal, and sequential information. U-time is based on U-Net that showed great performance of segmentation of biomedical images \cite{Ronneberger2015}. U-Net is composed of a series of contracting and expansive CNN without a fully connected layer. Since both models extracted better features than human-made features, sleep stage classification performance could be increased. However, deep learning models require vast data to optimize many parameters for each class. Collecting enough number of micro-sleep is inefficient because micro-sleep is a short and sudden event during driving. This inefficiency leads to a data imbalance problem that makes learning intricate \cite{Johnson2019} and decreases the performance of micro-sleep detection. 

In this study, we exploited night-sleep EEG, including wakefulness and non-rapid eye movement (NREM) sleep, to improve micro-sleep detection performance. We investigated that the model developed for sleep stage classification, including DSN and U-time, could detect micro-sleep. We hypothesized that micro-sleep corresponds to the early stage of NREM. To investigate the relationship between the micro-sleep and sleep stage, we used t-stochastic neighbor embedding (tSNE) with a spectrogram of EEG during sleep. Moreover, we proposed the deep learning method to improve the micro-sleep detection using night-sleep. Our results will help the understanding of micro-sleep and prevention of accidents from micro-sleep.

\section{Materials \& Methods}

\subsection{Datasets}
\subsubsection{Micro-sleep during Driving Simulation}
We used the validation set from track 2 in the 2020 International BCI Competition (KU-Driving, https://osf.io/pq7vb/). This dataset included a single EEG channel (Pz-Oz) from 10 subjects during the experiment. Experiments were performed with the driving simulator, HPRSS-RX-I (R-CRAFT, Korea). The driving road was a monotonous straight road in the virtual environment. During the experiment, there was no external stimulation except an alarm to wake the subjects when subjects leave the driving road. Participants reported sleepiness every five minutes using the Karolinska Sleepiness Scale (KSS), which is from 1 to 9 and indicates that the higher value means the more sleepy state of participants \cite{Gillberg2010}. When KSS was over seven, they considered the five minutes before report as micro-sleep.  

\subsubsection{Night-sleep at Home}

We used the Sleep-EDF from PysioNet \cite{Kemp2000, Goldberger2000}. The sleep-cassette subset of the Sleep-EDF consisted of 153 whole-night polysomnographic sleep recordings of healthy Caucasians age 25-101 taking no sleep-related medication. We selected the Pz-Oz EEG channel among polysomnography data. The expert sleep technician used the measured physiological signals to assign the sleep stages to 30 sec-long windows according to the Rechtschaffen and Kales (R\&K) manual \cite{Moser2009}. We used five sleep stages: Wake, NREM1, NREM2, NREM3, NREM4. We selected a total of 50 EEG data that were age from 25 to 56.

\subsection{Spectrogram of EEG and tSNE}

To investigate the relationship between micro-sleep and sleep stage, we obtained EEG spectrograms and compared them using tSNE. We obtained the $i$ th spectrogram from subject $j$ ($S_{ij}$) by following equation:
\begin{equation}
S_{ij}={\left | X_{ij}(\tau,\omega) \right |}^2\label{spectro}
\end{equation}
where $X(\tau,\omega)$ is the short-term Fourier transform of EEG signal, $x_{ij}(t)$. Therefore, 30 sec EEG signal ($x_{ij} \in \mathbb{R}^{3000}$) converted into spectrogram ($S_{i,j} \in \mathbb{R}^{33 \times 87}$).

To visualize the distribution of sleep stage and micro-sleep, we performed tSNE that reduced a dimension of data \cite{Maaten2008}. EEG signal and spectrogram are converted into embedded features ($\mathbb{R}^{2}$) according to the distance between data. We used the cosine function for distance measurement.

\begin{figure}[!tbp]
\centerline{\includegraphics{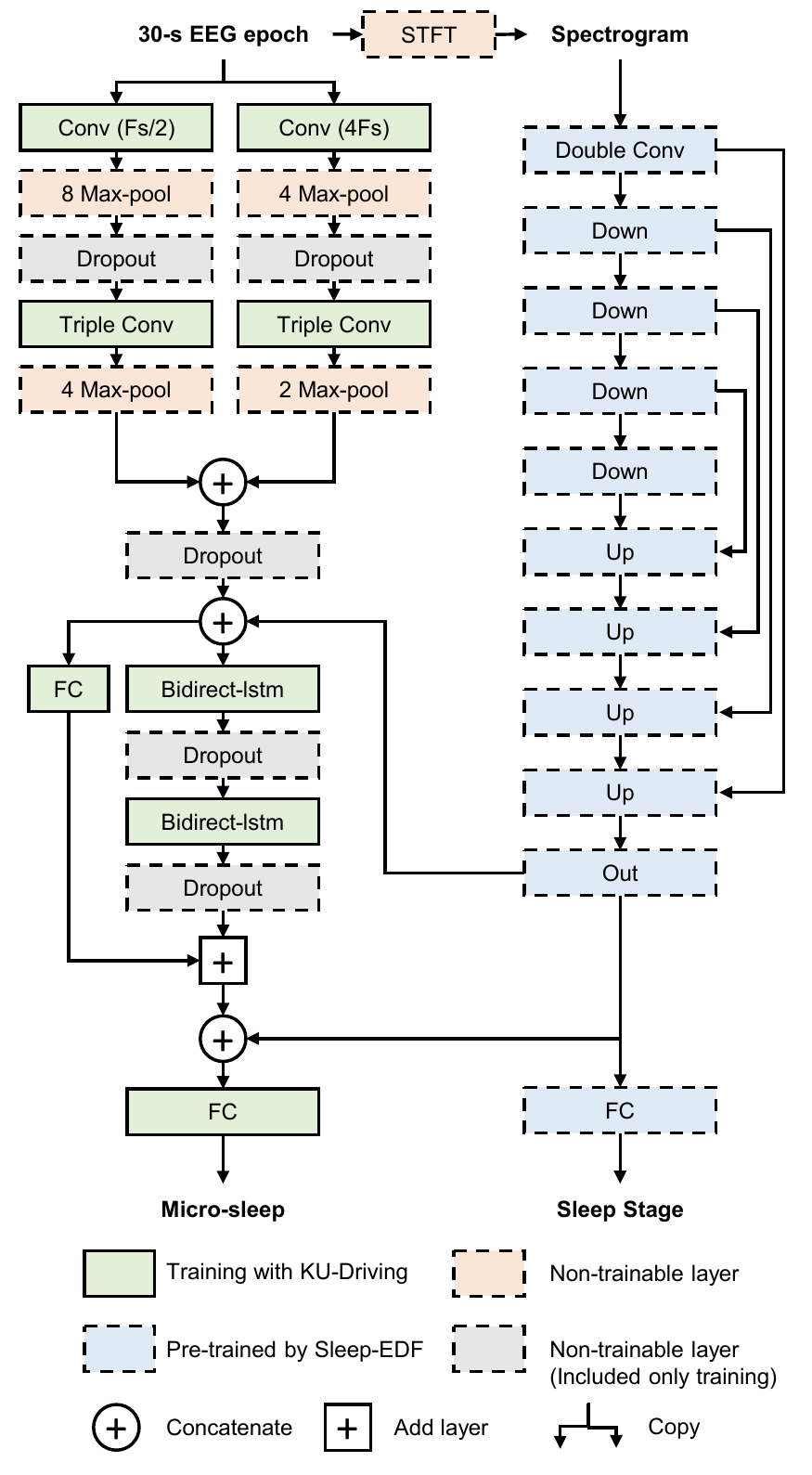}}
\caption{An overview architecture of DeepSleepNet \cite{Supratak2017} based micro-sleep detector and U-Net \cite{Ronneberger2015} based sleep stage extractor. The sleep stage extractor is pre-trained using Sleep-EDF and provides the sleep stage of a driver during driving. Micro-sleep detectors are the main model to detect micro-sleep during driving. Up and Down are modules to expand and contract feature maps, respectively. Short-term Fourier transform (STFT) was performed to provide a spectrogram. (${F}_{s}$: sampling frequency, FC: Fully-connected layer)}
\label{architecture}
\end{figure}

\subsection{Proposed Methods}
\subsubsection{U-Net for Sleep Stage Classification}

 We modified the U-Net to predict the sleep stage \cite{Ronneberger2015}. U-Net was designed for reconstructing data based on CNN with data contracting and expansive paths \cite{Ronneberger2015}. Spectrogram firstly entered into the contracting path. The contracting path was performed by a couple of 3$\times$3 convolutional layers followed by a rectified linear unit (ReLU) and 2$\times$2 max-pooling layer with stride 2 (Down in Fig. \ref{architecture}). The expansive path consisted of 2$\times$2 up-convolution, followed by concatenation with the cropped feature map from the contracting path and a couple of 3$\times$3 convolutions (Up in Fig. \ref{architecture}). The number of feature channels doubled at each stage of contraction but halved at the expansion path. To classify the sleep stage, we performed an out-convolutional layer followed by a fully connected layer (Out and FC in Fig. \ref{architecture}). We used the Adam optimizer with $10^{-3}$ learning rate and selected batches of size $128$. This model classified five classes: Wake, NREM1, NREM2, NREM3, and NREM4, to find out the sleep stage during driving.

\subsubsection{DSN and U-time for Micro-sleep Detection}
 
 We implemented the U-time for micro-sleep detection \cite{Perslev2019}. U-time is a modified version of U-Net by converting a 2-dimensional convolutional layer into a 1-dimensional convolutional layer. We used the Adam optimizer with $10^{-4}$ learning rate and $10^{-7}$ weight decay. We selected batches of size $8$.
 
 We also modified the DSN for micro-sleep detection (https://github.com/akaraspt/deepsleepnet)\cite{Supratak2017}. To capture the EEG signal's frequency and temporal features, two streams of 1-dimensional CNN with $F_{s}/2$ and $4F_{s}$-sized filters were designed ($F_{s}$: sampling frequency). After EEG was entered into each stream, each stream calculated frequency and temporal features. These were concatenated into the frequency-temporal features. To learn the sequential features of the sleep stage transition, frequency-temporal features were entered in the bidirectional LSTM. Sequential features were given through the calculation of bidirectional LSTM. In the end, sequential and frequency-temporal features were entered in the fully connected layer to detect the micro-sleep. DSN trained 1-dimensional CNN before training the bidirectional LSTM (Fig. \ref{architecture}). We skipped the over-sampling to avoid over-fitting. We used the Adam optimizer with $10^{-3}$ learning rate and $10^{-6}$ weight decay and selected batches of size $8$.
 
 \subsubsection{Micro-sleep Detector with Sleep Stage Extractor}
 
 We designed our model by combining a micro-sleep detector (U-time and DSN) and the sleep stage extractor (U-Net). Sleep stage extractor was pre-trained by Sleep-EDF without cross-validation. Spectrogram of KU-Driving entered into the pre-trained sleep stage extractor and obtained sleep stage features: 
 \begin{equation}
f_{ij}=SSE(S_{ij})\label{ssfeature}
\end{equation}
where SSE is the sleep stage extractor except for the fully connected layer. Micro-sleep detector concatenated the sleep stage features ($f_{ij}$) and the outcome of the model before the fully connected layer to inform the sleep stage during driving (Fig. \ref{architecture}). This was trained with the same hyper-parameters of the original model.

\begin{figure}[!tbp]
\centerline{\includegraphics{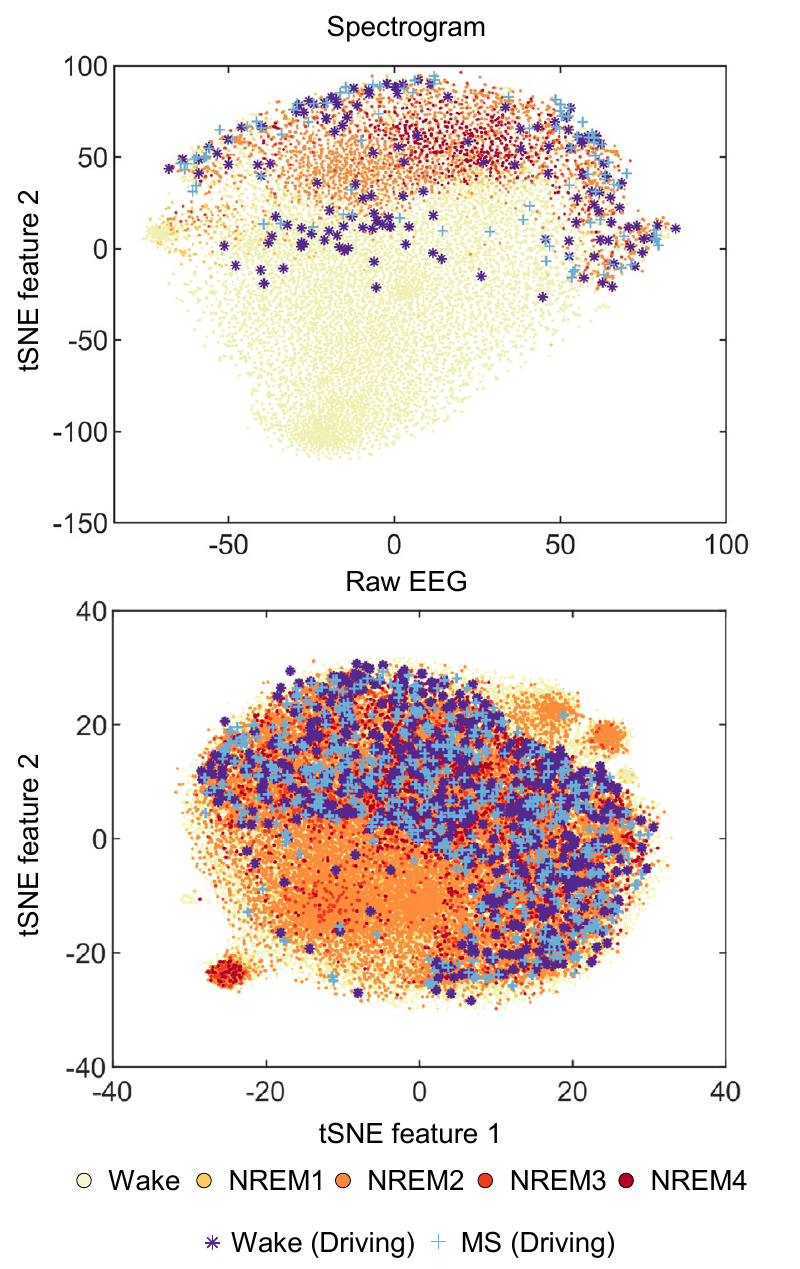}}
\caption{Distribution of spectrogram and raw EEG during night-sleep and micro-sleep. Dimension reduction was made by tSNE with a cosine distance function. MS means micro-sleep during driving.}
\label{tsne}
\end{figure}

\subsection{Evaluation and Metrics}

 We evaluated our model with the leave-one-subject-out approach to measure the generalization performance for a new subject. In both KU-Driving and Sleep-EDF datasets, we set one subject as a validation set in turn and the rest subjects as a training set. We measured the performance of our model by averaging all scores from each of the cross-validation folds. 
 
Our model's evaluation metric was Cohen's kappa coefficient, which is used for measuring reliability between two raters \cite{Cohen1960}. The following equation calculates the kappa:
\begin{equation}
    K \equiv {\frac{p_o-p_e}{1-p_e+\epsilon}} \label{kappa}
\end{equation}
where $p_o$ is the rate of agreement among raters, $p_e$ is the probability of agreement by chance, and $\epsilon$ is a small value to prevent the NaN value.

\begin{figure*}[!tbp]
\centerline{\includegraphics{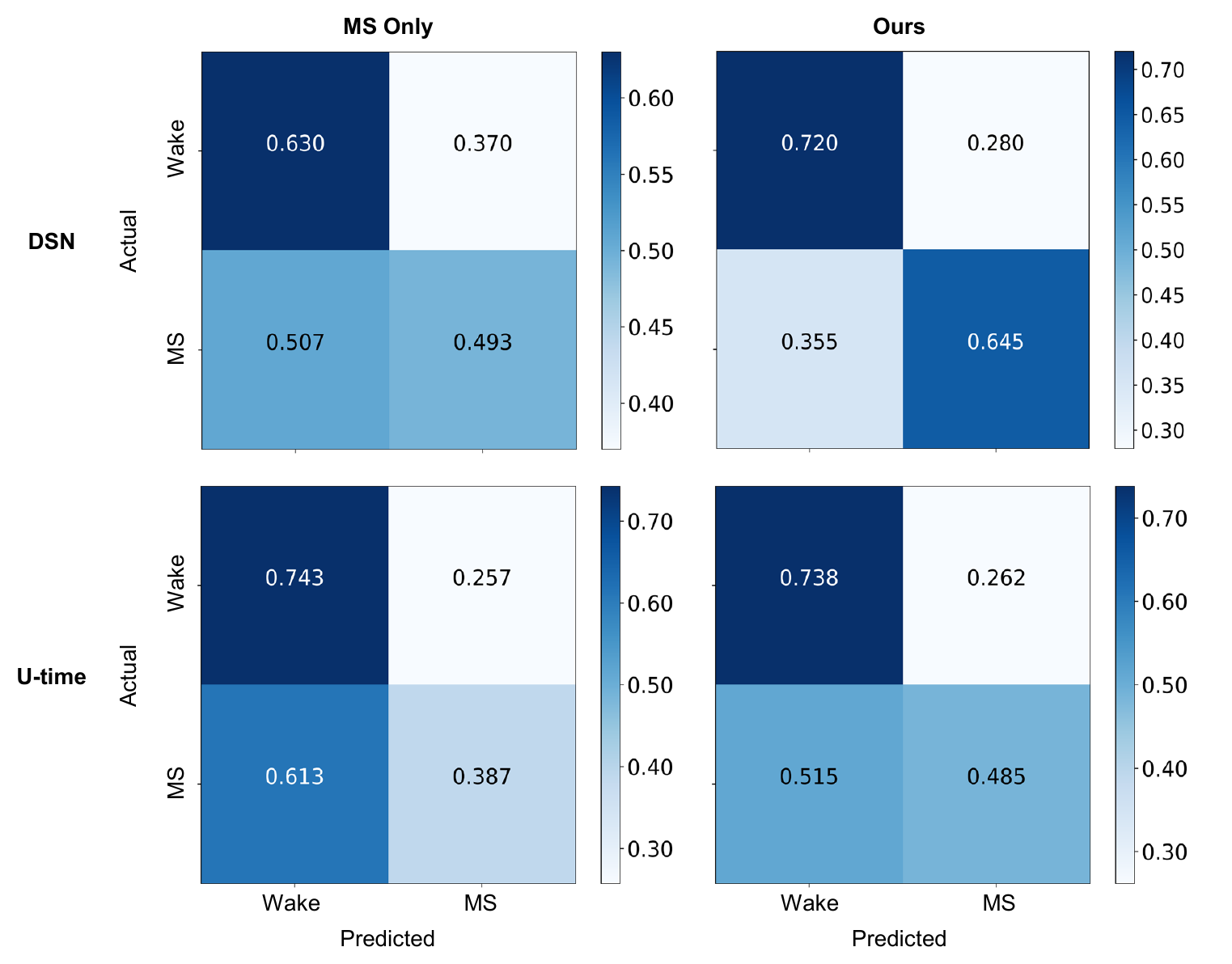}}
\caption{Confusion matrices of DSN and U-time according to training methods. The column represents training methods (left: using only KU-Driving, right: using KU-Driving and Sleep-EDF). Row indicates the prediction model (up: DSN, down: U-time). MS Only represents the model trained by only KU-Driving. Ours represents the model trained by KU-Driving and Sleep-EDF.}
\label{CM}
\end{figure*}

We also used a confusion matrix to compare the model's performance. The false alarm rate of micro-sleep was set at the rate that the model incorrectly predicted micro-sleep during wakefulness. Micro-sleep's true alarm rate was the ratio of trials that the model correctly predicted during actual micro-sleep. The missed micro-sleep was set at the rate that the model incorrectly predicted micro-sleep during actual micro-sleep.

\section{Results}

\subsection{Comparison of Distribution Between Night-sleep and Micro-sleep}

We found that raw EEG has an ambiguous distribution between sleep stages, but spectrogram has distinguishable distribution between sleep stages. When using raw EEG, trials spread around the origin regardless of the sleep stage. On the other hand, we observed a tendency to spread upward as the sleep deepened. Also, when the sleep stage was different, they were rarely located in one space (Fig. \ref{tsne}). 

Our results showed that micro-sleep is similar to NREM1 or NREM2. Wakefulness during driving overlapped with wakefulness during sleep in spectrogram distribution. Some trials labeled wakefulness during driving was located with micro-sleep. On the other hand, most trials labeled micro-sleep during driving overlapped with NREM1 or NREM2 (Fig. \ref{tsne}). 

\begin{table}[!tbp]
\caption{Performance of Deep Learning Model for Sleep-stage Classification and Micro-sleep Detection}

\begin{threeparttable}[hb]
    \begin{tabular}{llll}
    \hline
    \textbf{Training Dataset} & \textbf{Test Dataset} & \textbf{Model}      & \textbf{Kappa}      \\ \hline
    NS         & NS            & U-Net               & 0.723 ($\pm$ 0.137) \\
    MS         & MS            & U-time              & 0.127 ($\pm$ 0.068) \\
    MS         & MS            & DSN                 & 0.178 ($\pm$ 0.114) \\
    MS / NS     & MS            & Ours (U-time)       & 0.199 ($\pm$ 0.095) \\
    MS / NS     & MS            & Ours (DSN)          & 0.254 ($\pm$ 0.110) \\ \hline
    \end{tabular}

    \begin{tablenotes}
    \item[] MS and NS represent KU-Driving and Sleep-EDF, respectively.
    \end{tablenotes}
\end{threeparttable}
\end{table}

\subsection{Performance of Sleep Stage Classification}

We applied U-Net with fixed architecture to Sleep-EDF datasets. Table 1 represented the averaged metrics of each fold and the standard deviation of metrics. Sleep stage classification showed higher performance than detection of micro-sleep. Unlike U-time and DSN showed under 0.200 kappa, U-Net showed 0.723 kappa, which meant the model's classification was similar to an expert's classification.

\subsection{Performance of Micro-sleep Detection}

U-time and DSN were applied to KU-Driving datasets. U-time showed lower false alarm rate for micro-sleep than DSN (U-time: 0.257, DSN: 0.370), but DSN showed higher true alarm rate for micro-sleep than U-time (U-time: 0.387, DSN: 0.493, Fig. \ref{CM}).

With our approach using night-sleep, we observed that U-time and DSN performance were increased. Our approach improved the kappa of the basic model by 36\% and 30\% at U-time and DSN. Micro-sleep detectors showed a lower false alarm rate and a higher true alarm rate for micro-sleep than the traditional approach. In DSN, false alarm rate decreased (MS only: 0.370, Ours: 0.280) and true alarm rate increased (MS only: 0.493, Ours: 0.645). Moreover, true alarm rate increased (MS only: 0.387, Ours: 0.485), but false alarm rate slightly increased in U-time (MS only: 0.257, Ours: 0.262). Missed micro-sleep was also decreased in U-time (MS only: 0.613, Ours: 0.515). In DSN, we observed decrements of missed micro-sleep (MS only: 0.507, Ours: 0.355, Fig. \ref{CM}).

\section{Discussion}

Our results of distribution analysis provide the possibility of high similarity between sleep and NREM1 or NREM2. This consent with the previous study using fMRI and EEG \cite{Poudel2014, Jonmohamadi2016}. fMRI study showed that similar blood oxygen level-dependent signal occurred during NREM and micro-sleep \cite{Poudel2014}. Sleep spindle, which is EEG signatures of NREM2, was observed during micro-sleep \cite{Jonmohamadi2016}. Further research should investigate the functional connectivity, which was used to classify abnormal brain states \cite{Ding2013, PLoS2013, Chen2017}.

We demonstrated that the model developed for sleep stage classification could detect micro-sleep. DSN and U-time showed higher kappa than zero, which means the models' detection is better than random selection. Moreover, we only used a single EEG channel for practicality during a real driving situation, unlike the previous machine learning approach with eye movement information. It means micro-sleep detection is possible using only a single EEG channel. 

The sleep stage feature during driving is an important feature of detecting micro-sleep, even if the sleep stage is not objective and have inter-rater variability \cite{Perslev2019}. Recent studies revealed that a brain has more complex stages during sleep than five stages \cite{Stevner2019}. The sleep stage feature, which is the last layer before the fully connected layer and is used for weakly supervised object localization in the computer vision \cite{Zhou2016}, might represent a complex brain state during sleep. Further study, which investigates the neurophysiological evidence of the relationship between the sleep stage feature and complex brain states, will help understand human sleep.

Although we verified that U-time and DSN could predict micro-sleep, their performance is not acceptable for a real driving situation. Since DSN and U-time showed 0.37 and 0.51 F1 scores to detect NREM1 \cite{Supratak2017, Perslev2019}, they had the model's limitation to detect micro-sleep. Moreover, participants evaluated their state of sleep in the experiments of KU-Driving, which might occur many labeling noises. Due to the limitation of the model and labeling noises, we could not achieve acceptable performance to detect micro-sleep during driving. Therefore, we should develop a better model to detect NREM1 and collect data without labeling noise.

To prevent accidents caused by micro-sleep, we need to increase the true alarm rate. However, when the true alarm rate increased, the false alarm rate can also be increased like U-time. The false alarm might attract drivers' intentions and be another cause of an accident. We should increase the true alarm rate and decrease the false alarm rate simultaneously. When sleep stage features were applied to DSN, we observed the increment of the true alarm rate and decrements of false alarm rates. Therefore, we selected DSN with sleep stage features as the best model.

\section{Conclusion \& Future Works}

In conclusion, we demonstrated night-sleep EEG could improve the performance of micro-sleep detection, and micro-sleep is similar to NREM1 and NREM2. Moreover, we verified that the model developed for sleep stage classification could detect micro-sleep using only a single EEG channel. Our results will help the development of an alarming or preventing system during car driving and understanding of micro-sleep.

Further research will investigate the neurophysiological similarity of brain states during micro-sleep and night-sleep. Recently developed methods of weakly supervised object localization can be used to estimate more accurate brain states. Considering the neurophysiological similarity of brain states during micro-sleep and night-sleep, we will develop a better way to apply sleep stage features to a micro-sleep detector. Effective alarming system to prevent micro-sleep during car driving will be developed in the future.

\vspace{12pt}
\end{document}